\documentclass[wcp]{jmlr}

\usepackage{longtable}

\usepackage{booktabs}
\usepackage{amsmath}
\usepackage{amssymb}

\usepackage{multirow}
\usepackage{adjustbox}

\usepackage{afterpage}
\usepackage{booktabs}

\usepackage{hyperref}
\usepackage{kotex}

\newcommand{\nj}[1]{\textcolor{black}{#1}}
\newcommand{\ms}[1]{\textcolor{black}{#1}}

\title{Disentangling Options with Hellinger Distance Regularizer}

  \author{\Name{Minsung Hyun} \Email{minsung.hyun@snu.ac.kr}\\
   \Name{Junyoung Choi} \Email{djcola814@snu.ac.kr}\\
   \Name{Nojun Kwak} \Email{nojunk@snu.ac.kr}\\
   \addr Department of Transdisciplinary Studies\\
    Seoul National University\\
    Seoul, Korea Republic of.}

\begin{document}

\maketitle

\begin{abstract}
In reinforcement learning (RL), temporal abstraction \ms{still remains as} an important and unsolved problem. The options framework provided clues to temporal abstraction in the RL, and the option-critic architecture elegantly solved the two problems of finding options and learning RL agents in an end-to-end manner. However, it is necessary to examine whether the options learned through this method play a mutually exclusive role. In this paper, we propose a \textit{Hellinger distance} regularizer, a method for disentangling options. In addition, we will \ms{shed light on} various indicators from the statistical point of view to compare with the options learned through the existing option-critic architecture.
\end{abstract}
\begin{keywords}
Reinforcement learning, Deep learning, Temporal abstraction, Options framework
\end{keywords}

\section{Introduction}
\label{introduction}

Hierarchical learning has been considered as a major challenge among AI researchers \nj{in imitating} human beings. When solving a problem, human beings subdivide the problem to create each hierarchy and arrange it in a time sequence to unconsciously create an abstraction. For example, when we prepare dinner, we follow a series of complicated processes. The overall process of cooking the ingredients, cooking them along with the recipe and setting the food on the table includes detailed actions such as cutting and watering in detail. This \nj{process} can be described as a temporal approach and is currently being addressed in the study of AI \cite{minsky1961steps, fikes1972learning, kuipers1979commonsense, korf1983learning, iba1989heuristic, drescher1991made, dayan1993feudal, kaelbling1993hierarchical, thrun1995finding, parr1998reinforcement, dietterich1998maxq}. Temporal abstraction in reinforcement learning (RL) is a concept that \ms{a} temporally abstracted action is conceptualized to understand problems in a faster and efficient way by understanding problems hierarchically \nj{for efficient problem-solving.} 

However, how to implement temporal abstraction in RL is still an open question. The \nj{framework of options} \cite{sutton1999between} provided a basis for solving the RL problem from the viewpoint of \nj{temporal} abstraction through the semi-Markov Decision Process (sMDP) based on options. The options are \nj{defined as} temporally extended courses of \nj{actions}. \citet{mann2014scaling, mann2015approximate} showed that the model trained using the options converges faster than the model trained with the primitive \nj{actions}. \citet{bacon2017option} then proposed an option-critic architecture. In this architecture, they extracted features from deep neural networks and combined the actor-critic architecture with the options framework. With this approach, they solved two combined problems that finding options based on data without prior knowledge and training the RL agent \nj{to maximize reward} in an end-to-end manner. Later, \citet{harb2017waiting} devised a way to learn options that last a \ms{long} time by giving the cost of terminating the options. In \citet{harutyunyan2017learning}, they proposed a solution to the problem of degrading performance due to the sub-optimal options by training the termination function in an off-policy manner.

In this paper, we are going to analyze the model based on the option-critic architecture in different two perspectives and propose a method to improve the problems. First, we will verify that the options learned from option-critic plays the role of sub-policy. We expect each option to play a different role. In the option-critic architecture, the number of options is set as a hyperparameter without prior knowledge of the environment. In the process of optimizing the objective function, the \nj{different} learned options often follow a similar probability distribution \cite{bacon2017option}. It is certain that this is not the direction we expect for the options framework, and we can expect that similar options will lead to \nj{inefficient} learning.

Second, we will examine what options the agent \nj{learns} from a variety of perspectives. The previous studies have focused on creating a model that guarantees high reward and fast convergence based on confidence in the options framework \cite{vezhnevets2016strategic, bacon2017option, harb2017waiting, harutyunyan2017learning, vezhnevets2017feudal, tessler2017deep, smith2018inference}. However, it is also meaningful to look at how the learning \nj{progresses} as well as the performance.

The options \nj{can be represented} as stochastic probability distributions \nj{of actions}. So we will evaluate whether there is a significant difference between the learned options in terms of stochastic perspectives. First, we examine the probability distribution of each intra-option policy at the same state from the learned model. We then use statistical distance measures such as \textit{Kullback-Leibler divergence} \cite{kullback1951information} \nj{and} \textit{Hellinger distance} \cite{hellinger09}, which is a kind of \textit{f-divergence} \cite{csiszar2004information}. 
Also, the state space learned by the network can be expressed as t-SNE \cite{maaten2008visualizing} at the latent variable stage, and we will also check whether the options play different roles depending on how the different options are activated in the state space.

In order to overcome the problems of the existing option-critic architecture, we propose a way to induce the \nj{options} to learn mutually exclusive roles. Our approach is based on the assumption that intra-option policies would be mutually exclusive if they learn different probability distributions. We considered various statistical distances, and used \textit{Hellinger distance} as the regularizer that best matches our purpose. When we applied our method to Arcade Learning Environment (ALE) \cite{bellemare2013arcade} and MuJoCo \cite{todorov2012mujoco} environment, we could verify that the agent \nj{disentangles} the options better while maintaining performance.

To summarize, our contribution in this paper is as follows.
\begin{itemize}
    \item Propose a method of \textit{Hellinger distance} regularizer and show the possibility of distangling options in RL.
    \item Examine trained options by option-critic architecture and our methods in detail with diverse measures.
\end{itemize}

We will proceed this paper in the following order. First, we \nj{overview the} reinforcement learning and options, the background of the problem we want to solve. And we investigate how we can measure the difference between options in terms of statistical distance. Based on this, we propose a method to disentangle the options, and we show that disentangling options is possible through our proposed method from various viewpoints by conducting \nj{experiments} in ALE and MuJoCo environment.

\section{Background}
\label{background}

\subsection{Reinforcement Learning}
\label{rl}
In this paper, we work with Markov Decision Processes (MDPs), which are tuples $\langle S, A, r, P, \\\gamma \rangle$ \nj{consisting} of a set of \nj{states} $S$, a set of \nj{actions} $A$, a transition function $P_{a}(s, s')$, \nj{a} reward function $r : R_{a}(s, s')$, and \nj{a} discount factor $\gamma$. A policy $\pi$ is a behavior function determined by $Pr(s_{t+1} = s' | s_{t}=s, a_{t}=a)$ which indicates the probability of selecting \nj{an} action in \nj{a} given state. The value of a state $V^{\pi}(s)$ is defined as $V^{\pi}(s) = \mathbb{E}[\sum_{t=1}^{\infty}\gamma^{t}r_{t}]$ which indicates \nj{the} expected sum of discounted \nj{rewards} when following \nj{the} policy $\pi$.

\subsection{Options}
\label{options}

\textbf{The Options Framework}
\citet{sutton1999between} suggested the idea for learning temporal abstraction through \nj{an} extension of \nj{a} reinforcement learning framework based on \textit{semi-Markov decision process} (sMDP). 
They used the term \textit{options}, which include primitive actions as a special case. Any fixed set of options defines a discrete-time sMDP embedded within the original MDP. Options consist of three components: a policy $\pi : S \times A \xrightarrow{}[0, 1]$, a termination condition $\beta : S \xrightarrow{} [0, 1]$, and an initiation set $\mathbb{I} \subseteq S$. If \nj{an} option is chosen, then actions are selected through $\pi$ until \nj{the} termination term $\beta$ stochastically \nj{terminates an} option. They also suggest \textit{policies over options} $\mu: S \times O \xrightarrow{}[0,1]$ that selects an option $o \in O_{s_{t}}$ according to \ms{a} probability distribution $\mu(s_{t}, \cdot)$

\textbf{Option-Critic Architecture}
\citet{bacon2017option} \nj{suggested} \nj{an} end-to-end algorithm on learning \nj{an option}. They \nj{suggested} the \textit{call-and-return} option execution model, in which an agent picks option $\omega$ according to its policy over option $\pi_{\Omega}$, then follows the intra-option policy $\pi_{\omega}$ until termination $\beta_{\omega}$. \nj{The i}ntra-option policy $\pi_{\omega, \theta}$ of \nj{an} option $\omega$ is parameterized by $\theta$ and the termination function, $\beta_{\omega, \vartheta}$ of \nj{the} option $\omega$ is parameterized by $\vartheta$. They proved both intra-option policy and termination function are differentiable \nj{with respect to} their \nj{parameters} $\theta$ and $\vartheta$ so that they can design a stochastic gradient descent algorithm for learning options. The option-value function is defined as:

\begin{gather}
Q_{\Omega}(s, \omega) = \sum_{a}\pi_{\omega,\theta}(a|s)Q_{U}(s, \omega, a)
\label{option-value_function}
\end{gather}

where $Q_{U} : S \times \Omega \times A \xrightarrow{} \mathbb{R}$ is the value of executing an action in the context of state-option pair:

\begin{gather}
Q_{U}(s, \omega, a) = r(s, a) + \gamma\sum_{s'}P(s'|s,a)U(\omega, s')
\label{state-option-pair}
\end{gather}

\nj{Here, $U(\omega, s')$ is } \ms{the option-value function \textit{upon arrival}} \nj{of option $\omega$ at state $s'$ \cite{sutton1999between}.} In \citet{harb2017waiting}, they \nj{suggested} the deliberation cost function, which is essentially a negative reward that they give to the agent whenever it terminates an option.
\begin{equation}
\begin{split}
U(\omega, s') =& (1-\beta_{\omega, \vartheta}(s'))Q_{\Omega}(s', \omega) \\
  & +\beta_{\omega, \vartheta}(s')(V_{\Omega}(s')-c)
\label{deliberation_cost}
\end{split}
\end{equation}

Eq. \ref{deliberation_cost} explains the value function for \nj{an} option-critic architecture. $c$ indicates the deliberation cost which \nj{acts} like a regularizer presented in \citet{bacon2017option} but \nj{it takes a more significant role}.

\section{Distance between Options}
\label{distance}
In the \nj{o}ptions framework, the intra-option policy is expressed as a probability distribution \nj{over actions} when a state is given. We \nj{believe} the intra-option policy must show different probability distributions \nj{over actions} in order to argue that each option plays a different role in RL. Therefore, it is necessary to measure the distance between probability distributions to see whether the options are mutually exclusive. In this section, we look at \ms{a way to measure the distance between probability distributions in terms of statistical distance.} 


\textbf{Statistical distance}
\citet{csiszar1967information} defines the difference of two probability distributions $P, Q$ by the following \textit{f-divergence}.

\begin{gather}
    D_{f}(P \Vert Q) = \sum_{x} Q(x)f \left(\frac{P(x)}{Q(x)} \right) 
    \label{f_divergence}
\end{gather}

\nj{Here,} $f(t)$ is a convex function defined \nj{for} $t>0$ and $f(1)=0$ in Eq. \ref{f_divergence} \cite{csiszar2004information}. \nj{The} f-divergence is 0 when the probability distributions $P$ and $Q$ are equal and non-negative due to convexity. There are various instances of 
\textit{f-divergence} according to \nj{different choices of a} function $f$, and we used \textit{Kullback-Leibler divergence} and \textit{Hellinger distance} to measure the difference in options.

\textbf{Kullback-Leibler Divergence} (\nj{KLD}) \cite{kullback1951information} is a representative measure of the probability \ms{distributional} difference, and the function $f$ of \textit{f-divergence} is $f(t)= t \log t$. \nj{KLD} is defined as follows when the probability distributions $P$ and $Q$ are discrete and continuous, respectively.

\begin{gather}
D_{KL}(P \Vert Q) = -\sum_{x} P(x)\log\left(\frac{Q(x)}{P(x)} \right) \label{kl_disc} \\
D_{KL}(P \Vert Q) = -\int_{-\infty}^{\infty} p(x)\log\left(\frac{Q(x)}{P(x)} \right)dx \label{kl_cont} \\
\end{gather}

Since \nj{KLD} follows the properties of the \textit{f-divergence}, it is always non-negative and zero when the probability distributions $P$ and $Q$ are equal. But it is an asymmetric measure and the value $D_{KL}(P \Vert Q)$ is not equal to $D_{KL}(Q \Vert P)$. This is because \nj{KLD} can be expressed as the difference in the amount of additional information needed to reconstruct the probability distribution $P$ with probability distribution $Q$ \cite{cover2012elements}. In order to complement the asymmetry of \nj{KLD}, \citet{lin1991divergence} proposed {Jensen-Shannon divergence (\nj{JSD})}. However, we do not use \nj{JSD} because it is impossible to calculate a closed form solution for the continuous probability distribution and we cannot apply it to the continuous action space we would experiment with.

\textbf{Hellinger Distance} \nj{(HD)} \cite{hellinger09} is the case where the function $f$ of \textit{f-divergence} is $f(t)=1-\sqrt{t}$. And the discrete and continuous \textit{Hellinger distance} of the probability distributions $P$ and $Q$ are defined as follows.
\begin{gather}
H(P,Q) = \frac{1}{\sqrt{2}} \sqrt{\sum_{i=1}^{k}(\sqrt{P_{i}} - \sqrt{Q_{i}})^2}\label{hd_disc} \\
H(P,Q) = \sqrt{1-\int_{\infty}^{-\infty}\sqrt{P(x)Q(x)}dx}\label{hd_cont}
\end{gather}

In Eq. \ref{hd_disc} $P$ and $Q$ are $k$-dimensional multivariate distributions. The \ms{HD} can be interpreted as the L2-norm of the probability distribution\ms{s}. In addition, the \nj{HD} is bounded \nj{to} $[0, 1]$, while having the property of \textit{f-divergence}. The maximum distance of $H(P, Q)$ is achieved when the probability of $Q$ is zero for a set where the probability of $P$ is greater than 0, and vice versa.

\section{Disentangling Options}\label{distentangle_options}
\subsection{Hellinger Distance Regularizer}\label{hd_regularizer}
We \nj{believe} that the \nj{different} options must play a different role in the same RL environment in order to \nj{be meaningful} in terms of temporal abstraction. However, as we will check in the \nj{experiments} of Section \ref{experiments}, the options learned in the option-critic architecture may play a similar role. We \ms{, therefore,} propose a regularizer that can disentangle options from a statistical distance perspective. We will add the \ms{HD} to the loss and use it as a regularizer. As we have seen in Section \ref{distance}, the \ms{HD} can be calculated as a \nj{closed differentiable} form in both continuous and discrete probability distributions and \nj{has the desirable properties that it is non-negative and bounded in between 0 and 1.} 

On the other hand, the other candidate we looked at in Section \ref{distance} is possible to measure the difference in options, but \ms{it is} not suitable as \ms{a regularizer}. In the case of \ms{KLD}, it is sometimes used as a regularizer to narrow the difference between distributions like in the case of Variational Autoencoder \cite{kingma2013auto}. But \nj{it} is not appropriate as a regularizer for our case, because it does not have an upper bound while we \ms{intend} to widen the difference between distributions. If we use \ms{KLD} as a regularizer, the value will diverge to infinity and fail to train the model. 

We defined the \ms{HD} regularizer (hd-regularizer) as follows:
\begin{gather}
    L_{hd-reg} = \frac{1}{_{m}C_{2}}\sum_{P}\sum_{Q}H(P,Q),\quad P\neq Q
    \label{hd_reg}
\end{gather}

\nj{Here, $_aC_b$ is the number of $b$ combinations among $a$ choices} and the hd-regularizer loss calculates the average of the \textit{Hellinger distances} of two different options when there are $m$ options.

\begin{figure}[!t]
\centering
    \subfigure[OC]{\label{fig:oc_hist}
      \includegraphics[width=.45\linewidth]{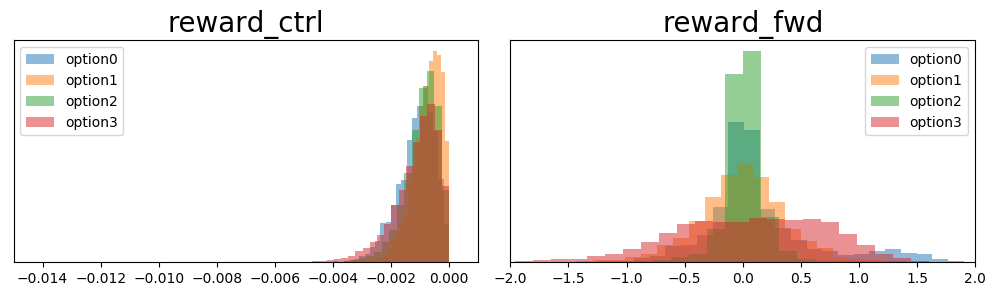}
    }
    \subfigure[OC+HD]{\label{fig:hd_hist}
      \includegraphics[width=.45\linewidth]{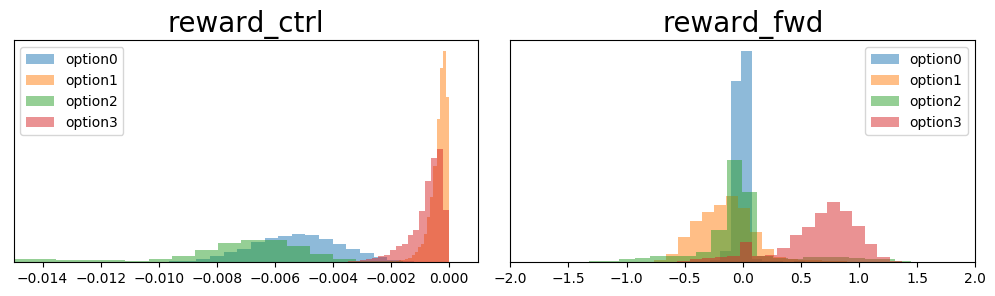}
    }
    \caption{Histogram of intrinsic rewards by options learned from (a) option-critic architecture (OC) and (b) our hd-regularizer method (OC+HD) in MuJoCo \textit{Swimmer-v2}. The horizontal axis is the corresponding reward. The distributions of different options in OC are largely overlapped.
On the other hand, options in OC+HD focus on different area of intrinsic rewards.} 
    \label{fig:swimmer_hist}
\end{figure}

\begin{table}[!t]
    \centering
    \renewcommand{\tabcolsep}{0.5mm}
    \vskip 0.15in
    \begin{center}
    \begin{small}
    \begin{sc}
    \adjustbox{max width=\linewidth}{
        \begin{tabular}{lcccc}
            \toprule
             $(\%)$ & o0 & o1 & o2 & o3\\
             \midrule
             OC & 0.75 & 2.37 & 55.62 & 41.25\\
             OC+HD & 5.75 & 59.87 & 9.12 & 25.25\\
             \bottomrule
        \end{tabular}}
    \end{sc}
    \end{small}
    \end{center}
    \vskip -0.1in
    \caption{Option use rate of option-critic architecture(OC) and hd-regularizer method(OC+HD) in MuJoCo \textit{Swimmer-v2}. Each column is an option label.}
    \label{tab:swimmer_option}
\end{table}

\subsection{Disentangling Options and Controllability}
\label{control}

We expect to add an hd-regularizer to the option-critic architecture to disentangle the intra-option policy that the option learns. We think that the expected utilization of \nj{disentangled} options is in the controlling \nj{of an} agent. The ultimate goal of RL is to create an agent that can perform \nj{like} \ms{a} human being or better, and the real world \nj{problems are} more complex than the \nj{environments} we are experimenting with. And in a complex environment, the reward to achieve must be more \nj{complex} \nj{than that of} a simple environment and \nj{be composed of multiple sub-rewards}. We think it would be possible to control the agent in the desired direction if the options are disentangled and \nj{different options perform} different functions focusing on different rewards, as we expected. It is possible to have control over an agent if the options are used differently, such as in a soccer match, where the coaches change aggressive or defensive tactics depending on the situation.

We try to check the feasibility of disentangling options and controllability based on experiments in environments with multiple rewards. \nj{When training an agent in the} MuJoCo's \textit{Swimmer-v2} environment, \nj{it has two implicit rewards whose combination makes up a total} \ms{reward}. Fig. \ref{fig:swimmer_hist} \nj{shows} the histogram of \nj{both intrinsic rewards for each option learned by the option-critic architecture (OC) and our method which uses a} \ms{hd-regularizer (OC+HD)}. In this figure, we have confirmed that when using our method, the options focus on different intrinsic rewards and \nj{we have} interpreted it as feasibility for controllability. See Section \ref{sub_reward} for more details.


\section{Experiments} \label{experiments}
Through experiments, we will identify the problems \nj{with} existing option-critic architectures and compare the aspects of our proposed hd-regularizer and \nj{the} option-critic architecture in various ways. We conduct experiments in the Arcade Learning Environment (ALE) \cite{bellemare2013arcade} and MuJoCo \cite{todorov2012mujoco} environment using the advantageous actor-critic (A2C) of \citet{schulman2017proximal} as the base algorithm. The network architecture and hyperparameters \nj{are also set equal to those of} the base algorithm for each environment. With regard to the option-critic architecture, we followed the structure of \citet{bacon2017option}. The number of options is fixed as 4 and the policy over option follows $\epsilon$-greedy policy \nj{with} $\epsilon=0.01$ and \nj{the} intra-option policy is trained by A2C method. We used RMSprop \cite{Tieleman2012} optimizer and updated the weight \nj{whenever 16 processes proceeded 5 steps}. The loss function consists of \nj{the} option-critic loss, \nj{the} entropy regularizer, and \nj{the} hd-regularizer. The option-critic loss consists of the policy gradient loss for the intra-option policy and value loss to estimate the option-value function, termination gradient loss and deliberation cost loss. We used an entropy regularizer to prevent intra-option policy from becoming deterministic too early and to encourage exploration \cite{williams1991function}. Finally, we added \nj{the} hd-regularizer to the loss after hyperparameter tuning. In order to prevent the intra-option policy from becoming fully deterministic in the discrete action space environment due to the hd-regularizer, we clamped the minimum probability of action of the policy to $10^{-4}$.
Details of the experimental setup are provided in the supplementary material.

\subsection{Arcade Learning Environment}\label{atari}
We experimented in the following six environments of Atari2600 provided by ALE with reference to \citet{bacon2017option, harb2017waiting}: \textit{AmidarNoFrameskip-v4, AsterixNoFrameskip-v4, BreakoutNoFrameskip-v4, HeroNoFrameskip-v4, MsPacmanNoFrameskip-v4}, \nj{and} \\ \textit{SeaquestNoFrameskip-v4}. Observation is a raw pixel image that stacked 4 frames \nj{as in} \citet{mnih2013playing}. The network consists of 3 convolutional layers with ReLU as the activation function and 1 fully-connected layer with option-value function, termination function, and intra-option policy stacked. And we trained $10^7$ steps per experiment. In addition, we used a clipped reward, the sign of actual reward, to reduce the performance impact of rewards with different scales for each environment.


\subsubsection{Reviewing Trained options} \label{ale_review}
First, through experiments, we have confirmed that temporal abstraction through the \nj{framework of options} is not always required for all RL problems. Table \ref{tab:ale_rewards} is the \nj{average reward and standard deviation of five runs of experiments} with different random seeds for each algorithm and Table \ref{tab:ale_option} summarizes the option use rate for each environment in one of \nj{the five} experiments. 
First, when comparing the results from a reward perspective, 
except for \textit{MsPacman}, all three methods show similar performance.
 In the case of \textit{MsPacman}, the performance of the option-critic architecture and our method, using the options framework, was better than the baseline. From this result, the temporal abstraction through options does not \ms{always} seem to guarantee performance improvement, in terms of rewards. 

\begin{table}[t]
    \begin{minipage}[t]{0.55\linewidth}
    \centering
    \vskip 0.15in
    \begin{center}
    \begin{small}
    \begin{sc}
    \adjustbox{max width=\linewidth}{
    \begin{tabular}{lccc} 
        \toprule
         Environment & A2C & OC & OC+HD \\
         \midrule
         Amidar & 346.81{\tiny$\pm$91.75} & 318.63{\tiny$\pm$51.73} & 305.85{\tiny$\pm$32.55} \\
         Asterix & 4412.80{\tiny$\pm$1318.23} & 4321.50{\tiny$\pm$1502.66} & 3985.70{\tiny$\pm$584.01} \\
         Breakout & 355.10{\tiny$\pm$38.31} & 349.54{\tiny$\pm$58.87} & 355.21{\tiny$\pm$25.25} \\
         Hero & 15834.35{\tiny$\pm$3459.75} & 14713.78{\tiny$\pm$2264.49} & 15229.56{\tiny$\pm$2640.07} \\
         MsPacman & 1361.18{\tiny$\pm$384.06} & 1934.50{\tiny$\pm$170.32} & 1910.96{\tiny$\pm$344.30} \\
         Seaquest & 1701.56{\tiny$\pm$50.85} & 1687.52{\tiny$\pm$49.64} & 1724.08{\tiny$\pm$25.31} \\
         \bottomrule
    \end{tabular}}
    \end{sc}
    \end{small}
    \end{center}
    \vskip -0.1in
        \caption{Average reward and its standard deviation from Arcade Learning Environment experiments. We tested with advantageous actor-critic(A2C), option-critic architecture(OC) and hd-regularization method(OC+HD). Each experiment was performed 5 times per environment. }
    \label{tab:ale_rewards}
    \end{minipage}
    \qquad
    \begin{minipage}[t]{0.44\linewidth}
        \centering
    \renewcommand{\tabcolsep}{0.5mm}
    \vskip 0.15in
    \begin{center}
    \begin{small}
    \begin{sc}
    \adjustbox{max width=\linewidth}{
        \begin{tabular}{l|cccc|cccc}
            \toprule
             $($\%$)$ & \multicolumn{4}{|c|}{OC} & \multicolumn{4}{|c}{OC+HD} \\
             Environment & o0 & o1 & o2 & o3 & o0 & o1 & o2 & o3 \\
             \midrule
             Amidar & 0.0 & 26.43 & 73.56 & 0.0 & 53.25 & 7.37 & 38.0 & 1.37 \\
             Asterix & 0.35 & 0.14 & 99.51 & 0.0 & 56.25 & 0.0 & 43.75 & 0.0 \\
             Breakout & 99.34 & 0.63 & 0.03 & 0.01 & 0.0 & 0.0 & 0.0 & 100 \\
             Hero & 0.0 & 1.23 & 29.61 & 67.16 & 0.25 & 0.0 & 24.5 & 75.25 \\
             MsPacman & 41.6 & 0.09 & 34.44 & 23.88 & 7.75 & 0.5 & 60.75 & 31.0\\
             Seaquest & 100 & 0.0 & 0.0 & 0.0 & 0.0 & 100 & 0.0 & 0.0 \\
             \bottomrule
        \end{tabular}}
    \end{sc}
    \end{small}
    \end{center}
    \vskip -0.1in
        \caption{Option use rate of option-critic architecture(OC) and hd-regularization method(OC+HD) in Arcade Learning Environment. Each column is an option label.}
    \label{tab:ale_option}
    \end{minipage}
\end{table}

We can also verify that the options framework is not always required through the use of trained options. Table \ref{tab:ale_option} summarizes the option use rate for each environment in one of our experiments. We can divide the environment into several groups. First, in \textit{Breakout} and \textit{Seaquest}, both the option-critic architecture and our method solves the problem using only one option. A possible explanation is that one option-value function would have an advantageous value for all states of the environment during $\epsilon$-greedy learning. In this case, the use of the options framework is not essential to \ms{solve RL problems.} The remaining four environments, except \textit{Breakout} and \textit{Seaquest}, belong to the second group and have been trained to use a variety of options in both methods. The second group can be said to have meaningful learning through the options framework, so we will continue to focus on these environments.

\begin{figure}[!t]
\centering
    \subfigure[OC]{
      \centering
      \includegraphics[width=.45\linewidth]{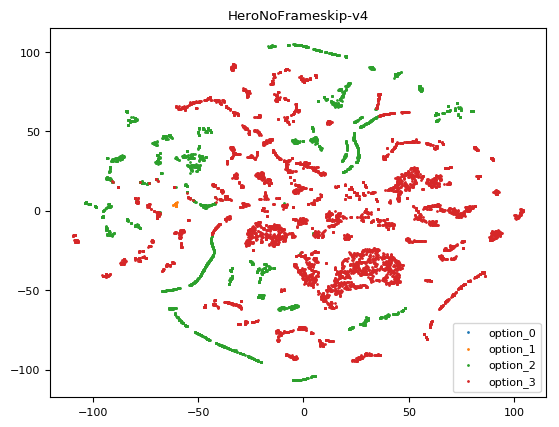}
      \label{fig:oc_hero_tsne}
    }
    \subfigure[OC+HD]{
      \centering
      \includegraphics[width=.45\linewidth]{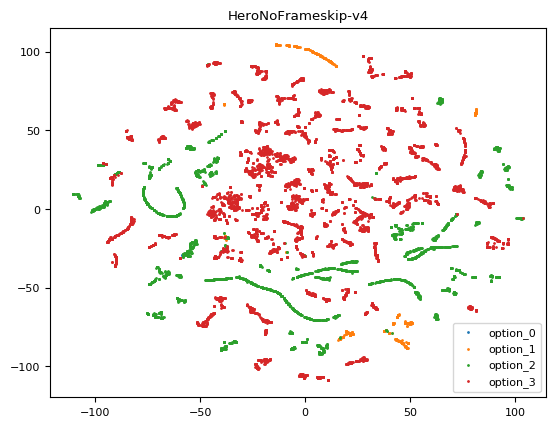}
      \label{fig:hd_hero_tsne}
    }
    \caption{t-SNE plot of option-critic architecture(OC) and hd-regularization method(OC+HD) in $Hero$. Each point represents a state and the color of the point is an option used in the state.}
    \label{fig:hero_oc_vs_hd_tsne}
\end{figure}
From Table \ref{tab:ale_option}, we can see that in Hero, both option-critic and our method use two major options, option2 and option3. Major options occupy more than 90 percent of the option usage, and minor options are rarely used. The state of two models can be examined in the latent variable space \ms{after} convolutional layers and one fully-connected layer explained in Section \ref{atari}. Fig. \ref{fig:hero_oc_vs_hd_tsne} is a representation of latent space with t-SNE \cite{maaten2008visualizing}, and both option-critic and our method appear to use options by disentangling states. If we look at this figure alone, it can be thought that both methods use the options well, but it does not \ms{if we} look at the options in more detail.
\begin{figure}[!t]
    \vskip 0.2in
    \begin{center}
    \centerline{\includegraphics[width=\textwidth]{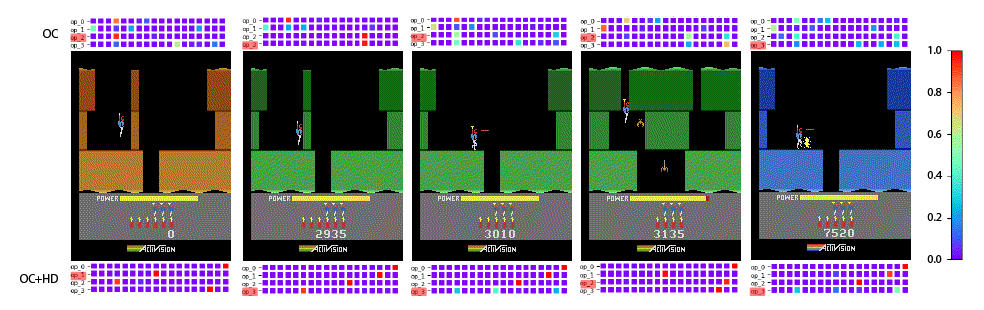}}
    \caption{Captured state images and intra-option policies of option-critic architecture(OC) and hd-regularization method(OC+HD) in \textit{Hero}. The heatmap above and below center the captured environment image represents a probability distribution of intra-option policy. Rows and columns of the heatmap are options and available actions. And the color of the heatmap is the probability of performing an action. An activated option at state is marked as red shaded.}
    \label{fig:hero_oc_vs_hd_dist}
    \end{center}
    \vskip -0.2in
\end{figure}
Fig. \ref{fig:hero_oc_vs_hd_dist} represents the probability distribution of the intra-option policy of the options of both methods when encountering a similar state in $Hero$. Top row is the intra-option policy of the option-critic architecture and bottom-row is the intra-option policy of our method with the hd-regularizer. The row of each intra-option policy is an option, column means discrete action, and the probability is expressed in a heatmap. And the shaded part in the option label of each row means that the option is activated at that state. In Table \ref{tab:ale_option}, option2 and option3 are major options of the option-critic model. In Fig. \ref{fig:hero_oc_vs_hd_dist}, all states of the intra-option policy learned in option-critic follows a similar probability distribution of option2 and option3. Using an option that follows a similar distribution means that both options have the same function and have done unnecessary learning. On the other hand, most of the hd-regularizer methods have different intra-option policies in the state\ms{s}.

\begin{table}[!t]
    \centering
    \renewcommand{\tabcolsep}{0.7mm}
    \vskip 0.15in
    \begin{center}
    \begin{small}
    \begin{sc}
    \adjustbox{max width=\linewidth}{
    \begin{tabular}{l|cc|cc}
        \toprule
          & \multicolumn{2}{|c|}{KLD} & \multicolumn{2}{|c}{HD} \\ 
         Environment & OC & OC+HD& OC & OC+HD \\ 
         \midrule
         Amidar & 1.22{\tiny$\pm$0.56} & 6.85{\tiny$\pm$1.71} & 0.43{\tiny$\pm$0.07} & 0.90{\tiny$\pm$0.08} \\ 
         
         Asterix & 1.98{\tiny$\pm$1.26} & 7.47{\tiny$\pm$1.44} & 0.50{\tiny$\pm$0.10} & 0.92{\tiny$\pm$0.05} \\ 
         
         Breakout & 2.78{\tiny$\pm$2.57} & 6.61{\tiny$\pm$2.87} & 0.53{\tiny$\pm$0.18} & 0.85{\tiny$\pm$0.85} \\ 
         
         Hero & 2.97{\tiny$\pm$2.02} & 7.21{\tiny$\pm$1.36} & 0.52{\tiny$\pm$0.05} & 0.93{\tiny$\pm$0.04} \\ 
         
         MsPacman & 10.64{\tiny$\pm$5.45} & 7.37{\tiny$\pm$1.16} & 0.78{\tiny$\pm$0.11} & 0.91{\tiny$\pm$0.07} \\ 
         
         Seaquest & 0.50{\tiny$\pm$0.33} & 7.09{\tiny$\pm$2.05} & 0.29{\tiny$\pm$0.08} & 0.93{\tiny$\pm$0.05} \\ 
         \bottomrule
    \end{tabular}}
    \end{sc}
    \end{small}
    \end{center}
    \vskip -0.1in
        \caption{An average distance between options and its standard deviation from option-critic architecture(OC) and hd-regularization method(OC+HD) measured by KLD and HD in Arcade Learning Environment.}
    \label{tab:ale_measure}
\end{table}

\subsubsection{Analysis with Measures} \label{ale_measure}
Here, we compare the similarities of the learned options numerically through \ms{two} measurement methods: \textit{kl-divergence} and \textit{Hellinger distance} discussed in Section \ref{background}. Table \ref{tab:ale_measure} is the average of the distance between options learned in each environment. In the case of \ms{KLD}, the values obtained by hd-regularizer method are larger than option-critic architecture in all environments except \textit{Hero}. And the \ms{HD} has larger values in our method for all environments and the values are close to 1 which is the upper bound of the HD. That is, the distance between the intra-option policy distributions is further with the hd-regularizer method. From this, we can see that the options learned by applying the hd-regularizer are disentangled in terms of statistical distance than those learned by the option-critic architecture. 

\subsection{MuJoCo Environment}\label{mujoco}
We also experimented for four environments of \ms{MuJoCo} with reference to \citet{smith2018inference}:\textit{Hopper-v2, Walker2D-v2, HalfCheetah-v2 \ms{and} Swimmer-v2}. \ms{MuJoCo} environment is based on continuous action space, so the environment setting is quite different from the previous setting. Observation is provided by \ms{MuJoCo} environment as state space and stacked 4 frames similar to ALE setting. We conducted the network structure of \citet{schulman2017proximal} and \citet{smith2018inference} with two hidden layers of 64 units with $tanh$ activation function for both value function and the policies. The intra-option policies were implemented as Gaussian distribution with hidden layers of mean and bias from the network. And we trained $2 \times 10^{6}$ steps for \ms{MuJoCo} environment.

\subsubsection{Reviewing Trained Option} \label{mujoco_review}
As we can see in Table \ref{tab:mujoco_rewards}, \ms{options work} more effectively in the continuous action space than in the \ms{ALE}. The result shows that the \ms{options are }
more effective than the environment with \ms{discrete actions} by showing a slightly higher or almost equal result in the option-critic architecture than the A2C. \ms{It can be said that the number of option is one factor that determine the complexity of the environment. And the policies of the two methods using the options framework here have bigger capacity than that of baseline method, A2C, because they have additional parallerized intra-option policy networks. Therefore, we think a learning method using the options framework will work better in a complex environment. From Table \ref{tab:mujoco_rewards}, } the complexity of a continuous action \ms{space} seems to maximize the benefit of the action capacity of the option. Especially, when using the option\ms{s} framework, there is no big difference in the action space of \textit{Swimmer}(2) and \textit{Hopper}(3), but the action space is wider and the complexity is higher in the environment of \textit{HalfCheetah}(6) and \textit{Walker2d}(6), we can see that the regularizing effect using \ms{the} options works effectively. This also applies to the assumption that the higher the complexity, the greater the effect of temporal abstraction.

We also confirmed that the option framework is not necessarily required in the \ms{MuJoCo} environment. In Table \ref{tab:mujoco_option}, we compared option use rates in each \ms{MuJoCo} environment. Like the \ms{ALE}, we can divide the experimented four environments into two groups. The first group is  \textit{Hopper} and \textit{HalfCheetah} environment which seem to be leaning towards one option-value function while learning with $\epsilon$-greedy. It tends to \ms{train a RL agent} by using only one option in the option-critic \ms{architecture}. In the case of using the hd-regularizer, \ms{the agent use more options in the \textit{Hopper}. This means that our method encourages the agent to learn more validate options by disentangling temporally abstracted options.}


\begin{table}[!t]
\centering
    \begin{minipage}[t]{0.50\linewidth}
    \centering
    \vskip 0.15in
    \begin{center}
    \begin{small}
    \begin{sc}
    \adjustbox{max width=\linewidth}{
    \begin{tabular}{lccc} 
        \toprule
         Environment & A2C & OC & OC+HD \\
         \midrule
         Hopper-v2 & 1299.3{\tiny$\pm$685.11} & 1393.2{\tiny$\pm$191.62} & 1377.7{\tiny$\pm$255.72} \\
         Walker2d-v2 & 1226.8{\tiny$\pm$805.98} & 1804.7{\tiny$\pm$785.08} & 1664.1{\tiny$\pm$374.32} \\
         HalfCheetah-v2 & 1421.6{\tiny$\pm$133.81} & 1403.6{\tiny$\pm$54.33} & 1773.1{\tiny$\pm$551.18} \\
         Swimmer-v2 & 34.73{\tiny$\pm$2.04} & 34.97{\tiny$\pm$1.61} & 36.13{\tiny$\pm$1.63} \\
         \bottomrule
    \end{tabular}}
    \end{sc}
    \end{small}
    \end{center}
    \vskip -0.1in
        \caption{Average reward and its standard deviation from MuJoCo Environment experiments. We tested with advantageous actor-critic(A2C), option-critic architecture(OC) and hd-regularization method(OC+HD). Each experiment was performed 5 times per environment.}
    \label{tab:mujoco_rewards}
    \end{minipage}
    \qquad
    \begin{minipage}[t]{0.42\linewidth}
        \centering
    \renewcommand{\tabcolsep}{0.5mm}
    \vskip 0.15in
    \begin{center}
    \begin{small}
    \begin{sc}
    \adjustbox{max width=\linewidth}{
        \begin{tabular}{l|cccc|cccc}
            \toprule
             $($\%$)$ & \multicolumn{4}{|c|}{OC} & \multicolumn{4}{|c}{OC+HD} \\
             Environment & o0 & o1 & o2 & o3 & o0 & o1 & o2 & o3 \\
             \midrule
             Hopper & 100 & 0.0 & 0.0 & 0.0 & 0.12 & 50.75 & 31.75 & 17.37 \\
             Walker2d & 4.12 & 12.75 & 3.75 & 79.37 & 10.75 & 10.25 & 75.12 & 3.88 \\
             HalfCheetah & 0.0 & 100 & 0.0 & 0.0 & 0.0 & 100 & 0.0 & 0.0 \\
             Swimmer & 88.37 & 11.5 & 0.12 & 0.0 & 5.00 & 85.25 & 3.62 & 6.12 \\

             \bottomrule
        \end{tabular}}
    \end{sc}
    \end{small}
    \end{center}
    \vskip -0.1in
        \caption{Option use rate of option-critic architecture(OC) and hd-regularization method(OC+HD) in MuJoCo Environment. Each column is an option label.}
    \label{tab:mujoco_option}
    \end{minipage}
\end{table}



\begin{figure}[!t]
\centering
    \subfigure[OC]{
      \centering
      \includegraphics[width=.45\linewidth]{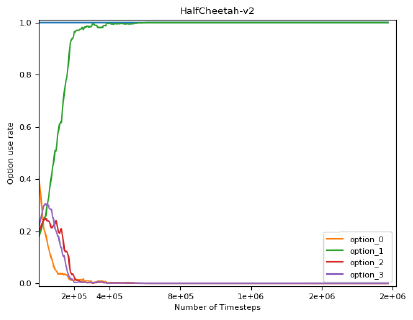}
      \label{fig:oc_halfcheetah_optionrate}
    }
    \subfigure[OC+HD]{
      \centering
      \includegraphics[width=.45\linewidth]{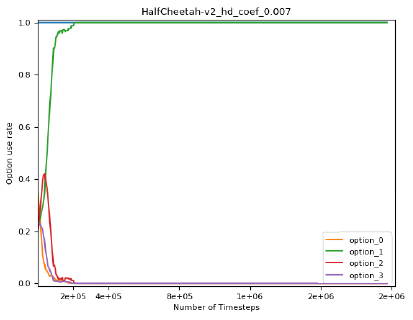}
      \label{fig:hd_halfcheetah_optionrate}
    }
    \caption{Training curves of option use rates for option-critic architecture(OC) and hd-regularization method(OC+HD) in $HalfCheetah$. The major option use rate of OC+HD converges to 100\% faster than that of OC.}
    \label{fig:optionrate_halfcheedah_ocvshd}
\end{figure}

The unusual point is that in \textit{HalfCheetah} environment, the final reward is higher than the A2C or option-critic method, although the proposed method concludes that it is much better to use only one option. Fig. \ref{fig:optionrate_halfcheedah_ocvshd} is a graph comparing the option use rates of the option-critic \ms{architecture} method and the proposed method.  Our method saturates quickly \ms{when} using only one option ($2 \times 10^5$ steps), whereas for the option-critic method, saturation is \ms{slower} ($8 \times 10^5$ steps). It seems that the difference in the rate of convergence increases the performance of \textit{HalfCheetah} in the proposed method.


\begin{figure}[ht]
\centering
    \subfigure[OC]{
      \centering
      \includegraphics[width=.45\linewidth]{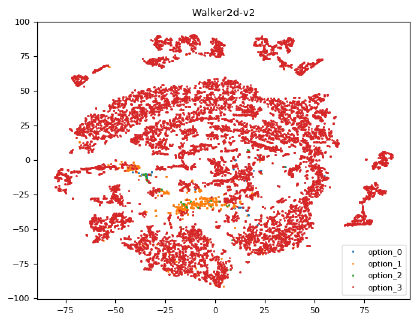}
      \label{fig:oc_walker_tsne}
    }
    \subfigure[OC+HD]{
      \centering
      \includegraphics[width=.45\linewidth]{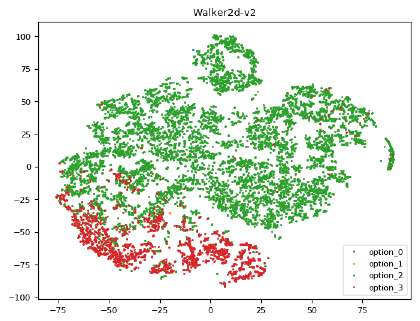}
      \label{fig:hd_walker_tsne}
    }
    \caption{t-SNE plot of option-critic architecture(OC) and hd-regularization method(OC+HD) in $Walker2d$. Each point represents a state and the color of the point is an option used in the state.}
    \label{fig:walker_oc_vs_hd_tsne}
\end{figure}

The second group learns to use more than one option, such as \textit{Walker2d} and \textit{Swimmer}. In this case, the option-critic architecture seems to play a sufficient role. However, as shown in Fig.\ref{fig:walker_oc_vs_hd_tsne}, the options learned in the option-critic method in the \textit{Walker2d} environment do not seem to be disentangled at all. Fig. \ref{fig:walker_oc_vs_hd_tsne} is a t-SNE analysis of \ms{the} options learned from the option-critic method and the proposed method. Respectively, t-SNE analysis allows you to see how each option is spread on each axis represented by the key features. In the case of the option-critic method, the network picked 2 major options but the options shown in the t-SNE \ms{plot} for the \textit{Walker2d} environment were not separated. However, in our method, we could visually confirm that each option was very disentangled. This suggests that each of the options in the proposed method can be interpreted as distributing roles by learning different functions compared to the option-critic method, which is advantageous for temporal abstraction.

\begin{table}[!t]
    \centering
    \renewcommand{\tabcolsep}{0.7mm}
    \vskip 0.15in
    \begin{center}
    \begin{small}
    \begin{sc}
    \adjustbox{max width=\linewidth}{
    \begin{tabular}{l|cc} 
        \toprule
           & \multicolumn{2}{|c}{HD} \\ 
         Environment & OC & OC+HD \\ 
         \midrule
         Hopper &  0.79{\tiny$\pm$0.12} & 0.99{\tiny$\pm$0.02} \\ 
         
         HalfCheetah &  0.84{\tiny$\pm$0.14} & 1.00{\tiny$\pm$0.002} \\ 
         
         Swimmer & 0.59{\tiny$\pm$0.12} & 0.99{\tiny$\pm$0.02} \\ 
         
         Walker2d &  0.92{\tiny$\pm$0.06} & 1.0{\tiny$\pm$0.004} \\ 
        
         \bottomrule
    \end{tabular}}
    \end{sc}
    \end{small}
    \end{center}
    \vskip -0.1in
    \caption{An average distance between options and its standard deviation from option-critic architecture(OC) and hd-regularization method(OC+HD) measured by HD in MuJoCo Environment.}
    \label{tab:mujoco_measure}
\end{table}

\subsubsection{Analysis with Measure} \label{mujoco_measure}
We tried to compare \ms{HD} as well as various measures in \ms{MuJoCo} environment. However, \ms{in the case of KLD which used in the \ms{ALE}, KLD has the value of zero or infinity when the two probability distributions are entirely exclusive. }
For the HD, as in Table. \ref{tab:mujoco_measure}, the proposed method had a value close to 1 in all \ms{environments}. option-critic method had a high value in environments except \textit{Swimmer}, but it did not record a higher distance than the proposed method.


\subsection{Disentangling Options by Intrinsic Reward}
\label{sub_reward}
As mentioned in Section \ref{control}, we can see the feasibility of using disentangled options for controlling agent by how each option is activated for intrinsic rewards. So we tested it in environments with multiple intrinsic rewards. The reward of \textit{Swimmer-v2} environment in MuJoCo implicitly consists of a control reward and a forwarding reward. The control reward is the reward for minimizing the amount of control and is inversely proportional to the absolute value of the action, and the forwarding reward is proportional to the amount of position change associated with moving the agent. In this case, as described in Table \ref{tab:swimmer_option}, the option-critic architecture uses option 1 and option 2 as major options, and our method uses option 0 and option 2 as major options. Fig. \ref{fig:swimmer_hist} is the histogram of the intrinsic rewards when each option is activated. In Fig. \ref{fig:swimmer_hist}(a), the option-critic architecture, the control reward is similar in all options. And the forwarding reward has the same locality around 0 despite different variance between major options. Fig. \ref{fig:swimmer_hist}(b) is the histogram of the model using the hd-regularizer. In this case, the control reward shows a similar frequency distribution to the option 0 and 2, while the forwarding reward shows different frequency distributions. In other words, the option 2 is considered to be an option optimized for maximizing the forwarding reward. From this, we can see that the options function differently in the reward aspect due to the hd-regularizer.


\section{Conclusion} \label{conclusion}
We have proposed an hd-regularizer that can disentangle the options. Through experiments, we compared and analyzed the \ms{learned} options from the existing method and the proposed method in various perspectives. As a result, we confirmed that the proposed hd-regularizer method disentangles the intra-option policies better than the option-critic architecture from the statistical distance perspective.

Although we somehow succeeded in disentangling the options, it did not succeed in interpreting the meaning of the options represented by the non-linear approximation. In order for the options framework to have an advantage, it is important to learn the temporal abstraction representation through which it can be intuitively understood. Controlling an agent would be possible in the direction we want, based on the options supported by the interpretation. Also, the options framework, including our method, did not perform well in all RL environments. This may be due to an environment in which temporal abstraction is not essential. However, a good algorithm should be able to cope with such an environment robustly. In order to continue the study of temporal abstraction in the future, it is necessary to concentrate on the methodology that works robustly in any environment.


\bibliography{acml19}
\end{document}


\maketitle

\section{Experiment Settings}

\subsection{Arcade Learning Environment}
\subsubsection{Network Structure}

\begin{figure*}[!h]
    \vskip 0.2in
    \begin{center}
    \centerline{\includegraphics[width=\textwidth]{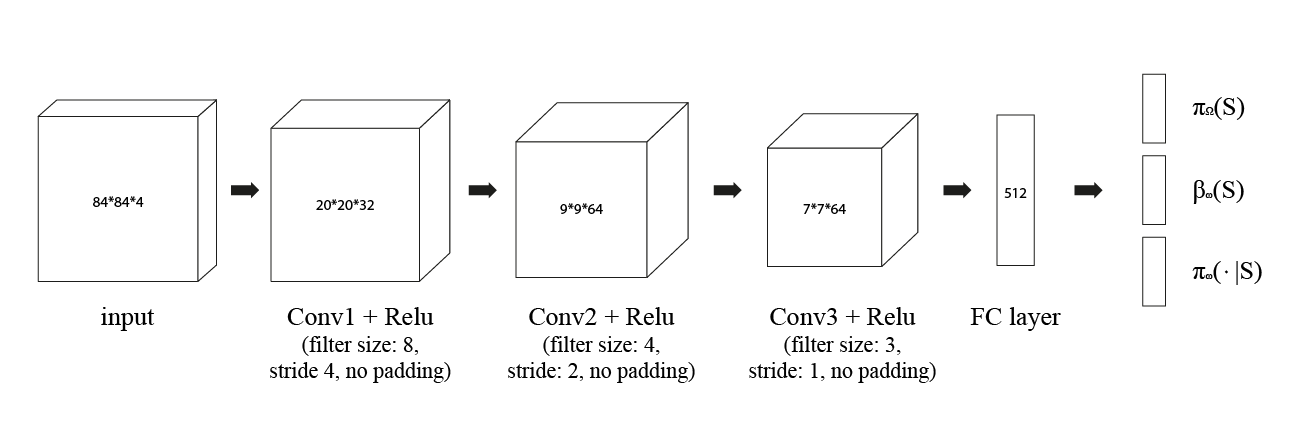}}
    \caption{Network Structure for Arcade Learning Environment}
    \label{fig:network_ale}
    \end{center}
    \vskip -0.2in
\end{figure*}

\subsubsection{Hyperparameters}
\begin{table}[!h]
    \centering
    \begin{tabular}{l|c}
         Hyperparameter & Value \\
         \midrule
         Learning rate & 0.0007 \\
         RMSprop smoothing constant & 0.99 \\
         Value loss coef. & 0.5 \\
         Entropy loss coef. & 0.01 \\
         Deliberation cost coef. & 0.01 \\
         hd-regularizer loss coef. & 0.007 \\
         $\epsilon$ of policy over options & 0.01 \\
         Number of processes & 16 \\
         Number of steps for update & 5 \\
         Training steps & $10^7$
    \end{tabular}
    \caption{Hyperparamerters of experiment in Arcade Learning Environment.}
    \label{tab:ale_hyperpara}
\end{table}
\newpage

\subsection{MuJoCo Environment}
\subsubsection{Network Structure}
\begin{figure*}[!h]
    \vskip 0.2in
    \begin{center}
    \centerline{\includegraphics[width=\textwidth]{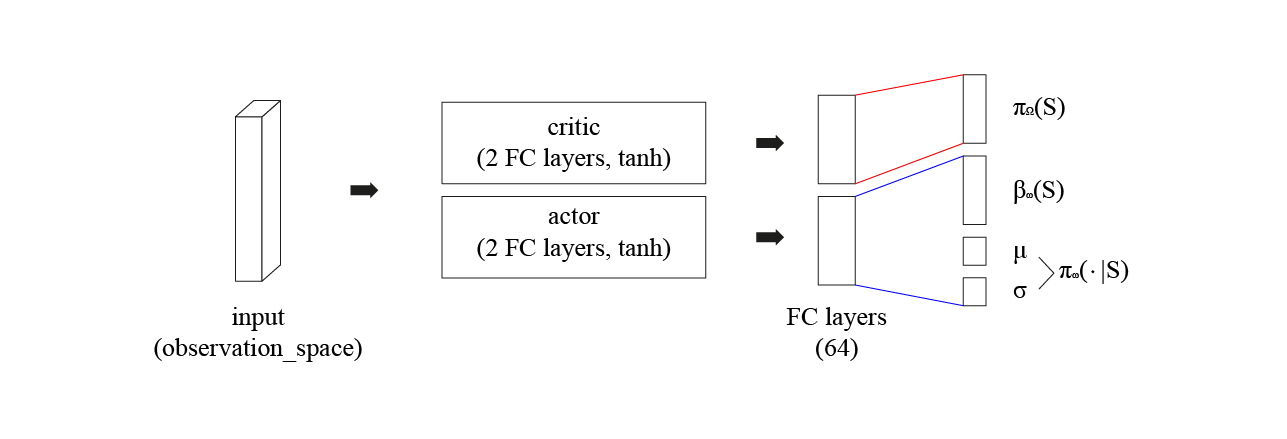}}
    \caption{Network Structure for MuJoCo environment}
    \label{fig:network_mujoco}
    \end{center}
    \vskip -0.2in
\end{figure*}

\subsubsection{Hyperparameters}
\begin{table}[!h]
    \centering
    \begin{tabular}{l|c}
         Hyperparameter & Value \\
         \midrule
         Learning rate & 0.0003 \\
         RMSprop smoothing constant & 0.99 \\
         Value loss coef. & 0.5 \\
         Entropy loss coef. & 0.0001 \\
         Deliberation cost coef. & 0.01 \\
         hd-regularizer loss coef. & 0.007 \\
         $\epsilon$ of policy over options & 0.01 \\
         Number of processes & 16 \\
         Number of steps for update & 5 \\
         Training steps & $2 \times 10^6$
    \end{tabular}
    \caption{Hyperparamerters of experiment in MuJoCo Environment.}
    \label{tab:mujoco_hyperpara}
\end{table}

\newpage

\section{Training Curves}
\subsection{Arcade Learning Environment}
\begin{figure}[h]
    \centering
    \subfigure[$Amidar$]{
        \centering
        \includegraphics[width=0.33\linewidth]{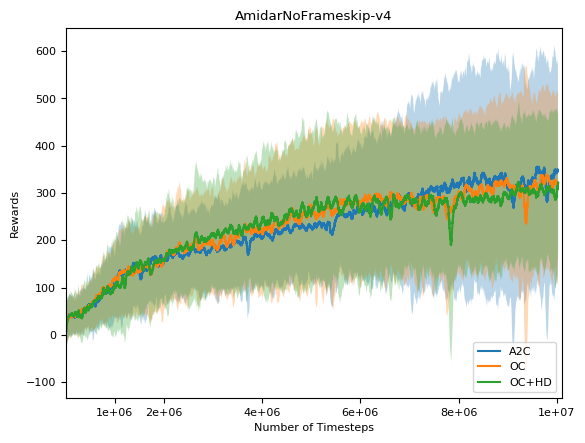}
        \label{fig:amidar_reward}
    }%
    \subfigure[$Asterix$]{
        \centering
        \includegraphics[width=0.33\linewidth]{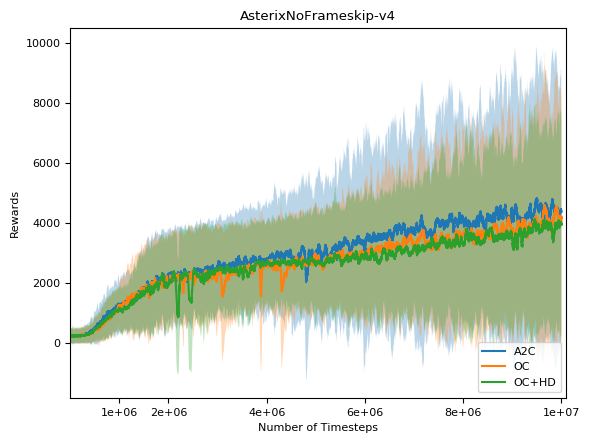}
        \label{fig:asterix_reward}
    }%
    \subfigure[$Breakout$]{
        \centering
        \includegraphics[width=0.33\linewidth]{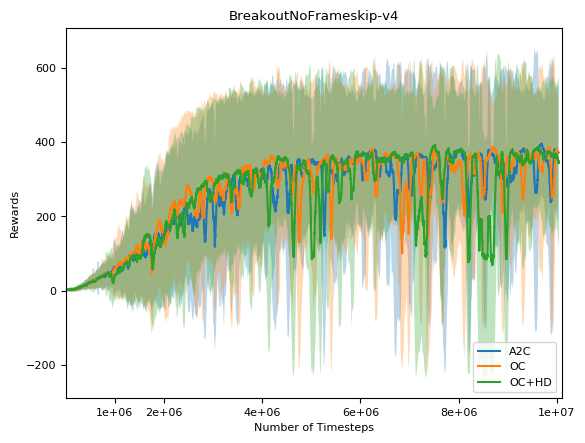}
        \label{fig:breakout_reward}
    }%
    \vskip\baselineskip
    \subfigure[$Hero$]{
        \centering
        \includegraphics[width=0.33\linewidth]{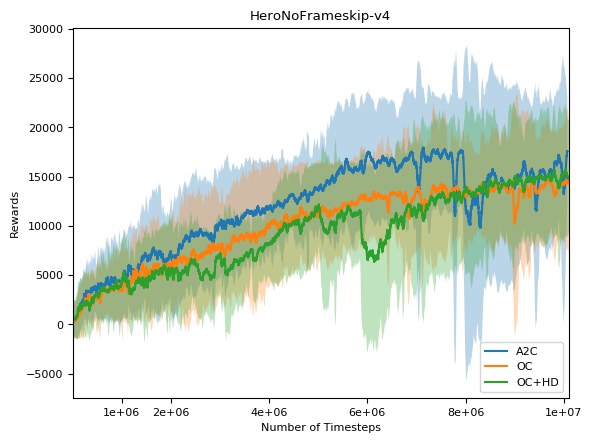}
        \label{fig:hero_reward}
    }%
    \subfigure[$MsPacman$]{
        \centering
        \includegraphics[width=0.33\linewidth]{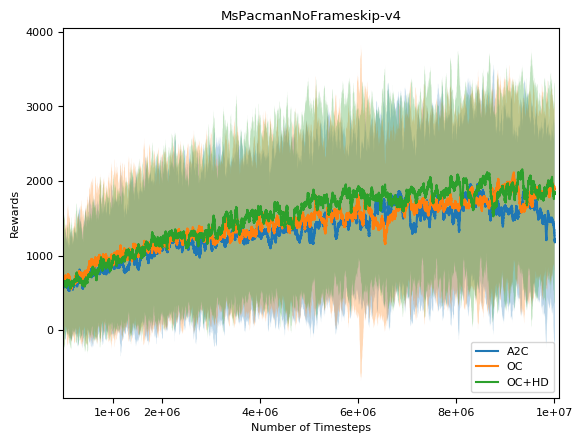}
        \label{fig:mspacman_reward}
    }%
    \subfigure[$Seaqeust$]{
        \centering
        \includegraphics[width=0.33\linewidth]{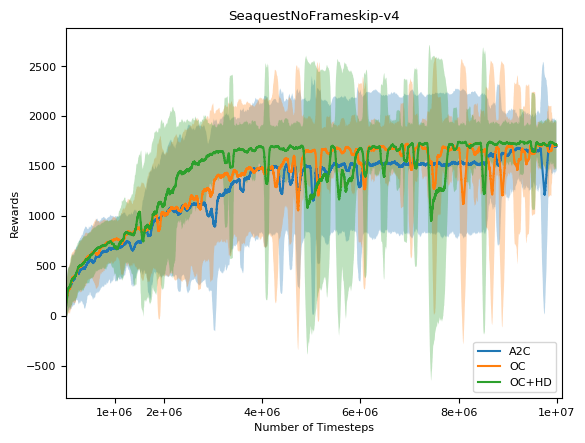}
        \label{fig:seaqeust_reward}
    }%
    \caption{Rewards training curves in Arcade Learning Environment. Bold straight line is moving average of 200 episodes and shaded area means $95\%$ confidence interval.}
    \label{fig:comb_reward}
\end{figure}

\begin{figure}[!h]
    \centering
    \subfigure[\textit{Amidar}]{
        \centering
        \includegraphics[width=0.33\linewidth]{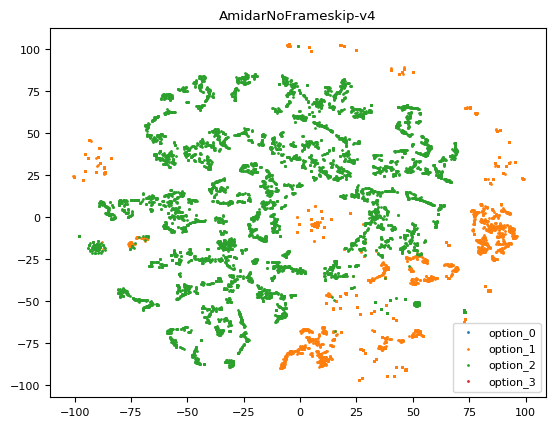}
        \label{fig:amidar_oc_tsne}
    }%
    \subfigure[\textit{Asterix}]{
        \centering
        \includegraphics[width=0.33\linewidth]{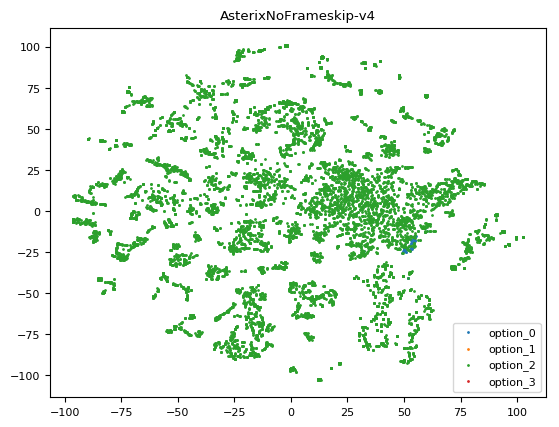}
        \label{fig:asterix_oc_tsne}
    }%
    \subfigure[\textit{Breakout}]{
        \centering
        \includegraphics[width=0.33\linewidth]{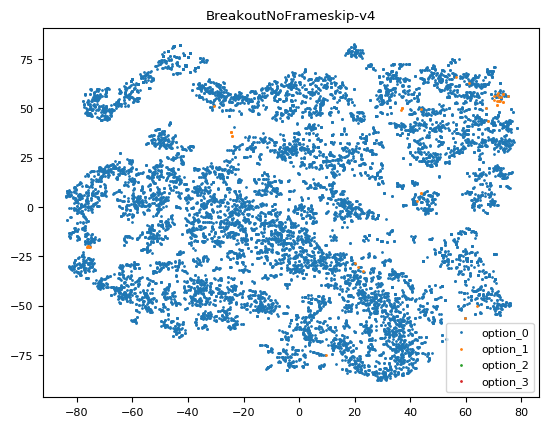}
        \label{fig:breakout_oc_tsne}
    }%
    \vskip\baselineskip
    \subfigure[\textit{Hero}]{
        \centering
        \includegraphics[width=0.33\linewidth]{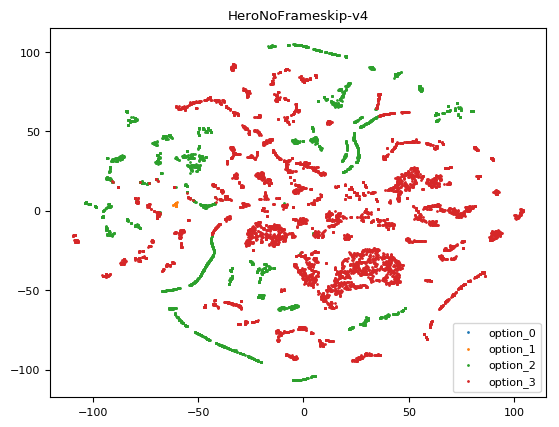}
        \label{fig:hero_oc_tsne}
    }%
    \subfigure[\textit{MsPacman}]{
        \centering
        \includegraphics[width=0.33\linewidth]{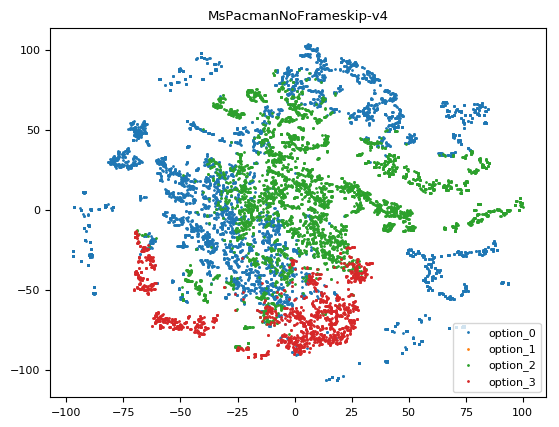}
        \label{fig:mspacman_oc_tsne}
    }%
    \subfigure[\textit{Seaqeust}]{
        \centering
        \includegraphics[width=0.33\linewidth]{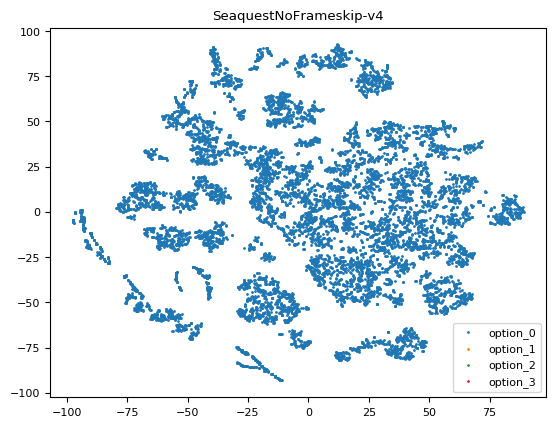}
        \label{fig:seaqeust_oc_tsne}
    }%
    \caption{t-SNE plot of the model trained with the option-critic architecture in Arcade Learning Environment.}
    \label{fig:comb_oc_tsne}
\end{figure}

\begin{figure}[!h]
    \centering
    \subfigure[\textit{Amidar}]{
        \centering
        \includegraphics[width=0.33\linewidth]{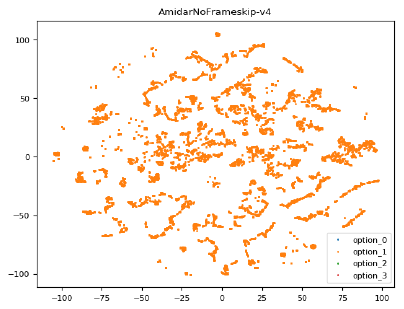}
        \label{fig:amidar_hd_tsne}
    }%
    \subfigure[\textit{Asterix}]{
        \centering
        \includegraphics[width=0.33\linewidth]{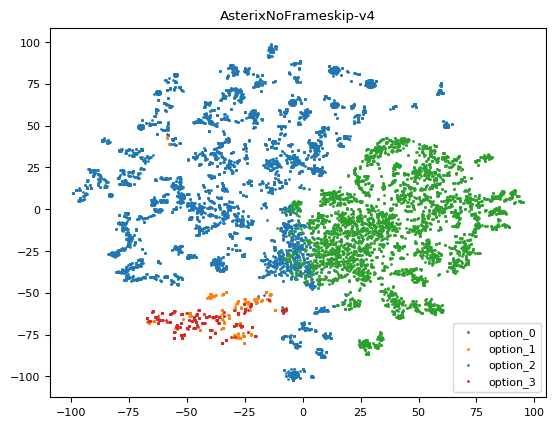}
        \label{fig:asterix_hd_tsne}
    }%
    \subfigure[\textit{Breakout}]{
        \centering
        \includegraphics[width=0.33\linewidth]{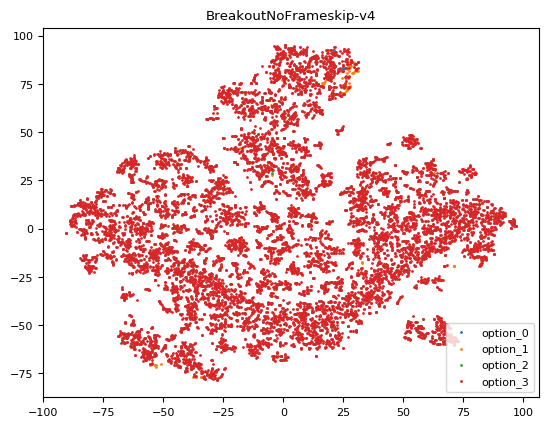}
        \label{fig:breakout_hd_tsne}
    }%
    \vskip\baselineskip
    \subfigure[\textit{Hero}]{
        \centering
        \includegraphics[width=0.33\linewidth]{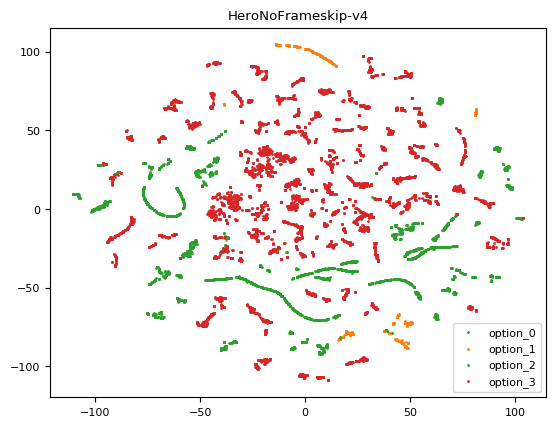}
        \label{fig:hero_hd_tsne}
    }%
    \subfigure[\textit{MsPacman}]{
        \centering
        \includegraphics[width=0.33\linewidth]{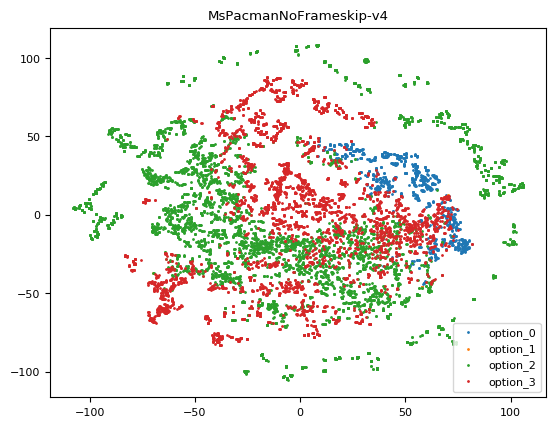}
        \label{fig:mspacman_hd_tsne}
    }%
    \subfigure[\textit{Seaqeust}]{
        \centering
        \includegraphics[width=0.33\linewidth]{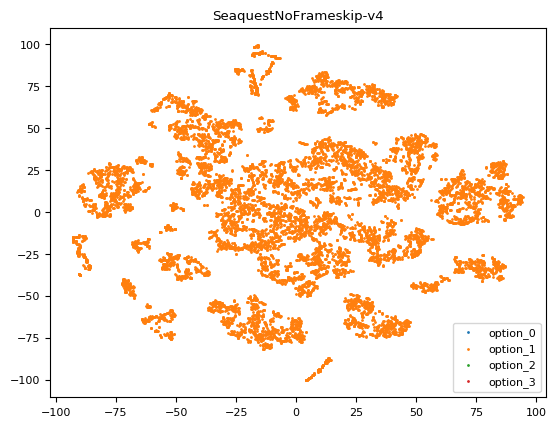}
        \label{fig:seaqeust_hd_tsne}
    }%
    \caption{t-SNE plot of the model trained with the hd-regularizer method in Arcade Learning Environment.}
    \label{fig:comb_hd_tsne}
\end{figure}
\phantom{this is for blank section}

\newpage

\subsection{MuJoCo Environment}
\begin{figure}[!h]
    \centering
    \subfigure[\textit{HalfCheetah}]{
        \centering
        \includegraphics[width=0.33\linewidth]{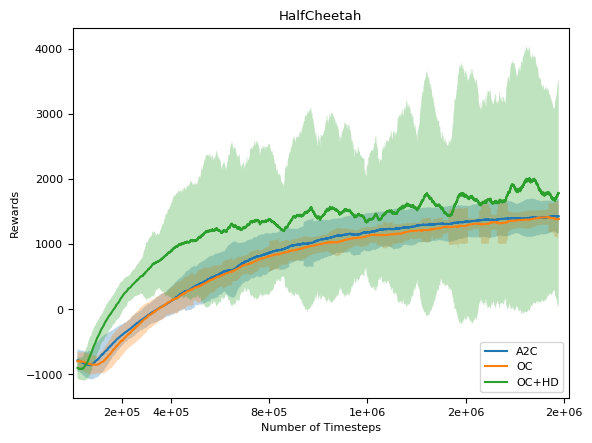}
        \label{fig:Halfcheetah_training_curve}
    }%
    \subfigure[\textit{Hopper}]{
        \centering
        \includegraphics[width=0.33\linewidth]{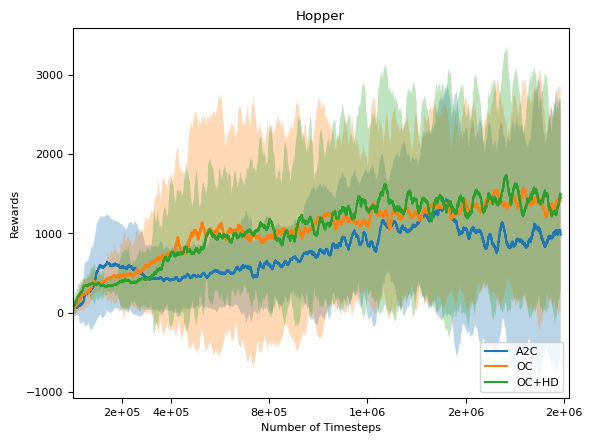}
        \label{fig:Hopper_training_curve}
    }%
    \vskip\baselineskip
    \subfigure[\textit{Swimmer}]{
        \centering
        \includegraphics[width=0.33\linewidth]{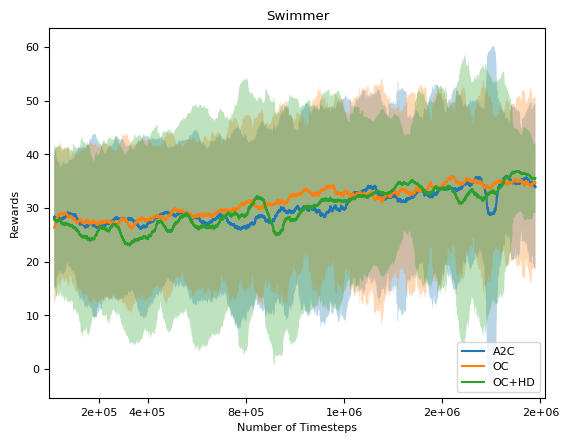}
        \label{fig:Swimmer_training_curve}
    }%
    \subfigure[\textit{Walker2D}]{
        \centering
        \includegraphics[width=0.33\linewidth]{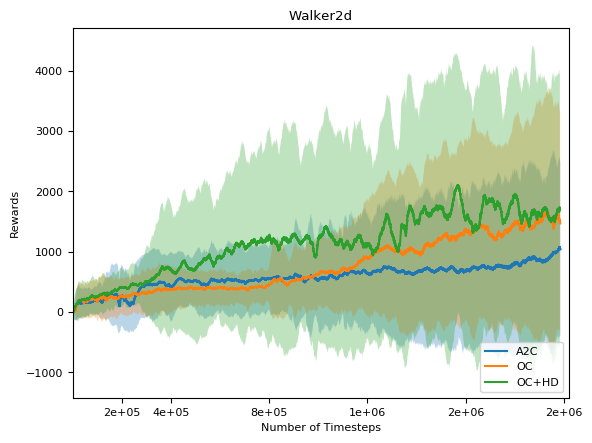}
        \label{fig:Walker2d_training_curve}
    }%
    \caption{Rewards training curves in MuJoCo environemnt. Bold straight line is moving average of 200 episodes and shaded area means $95\%$ confidence interval.}
    \label{fig:mujoco_training_curve}
\end{figure}

\begin{figure}[!h]
    \centering
    \subfigure[\textit{HalfCheetah}]{
        \centering
        \includegraphics[width=0.33\linewidth]{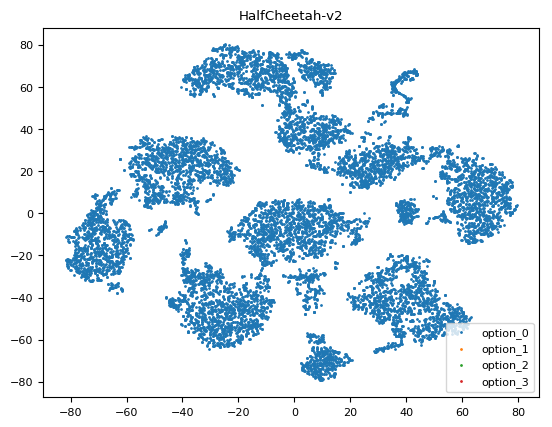}
        \label{fig:Halfcheetah_oc_tsne}
    }%
    \subfigure[\textit{Hopper}]{
        \centering
        \includegraphics[width=0.33\linewidth]{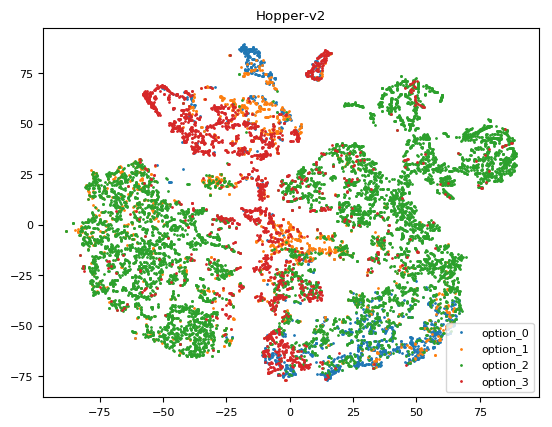}
        \label{fig:Hopper_oc_tsne}
    }%
 
    \vskip\baselineskip
    \subfigure[\textit{Swimmer}]{
        \centering
        \includegraphics[width=0.33\linewidth]{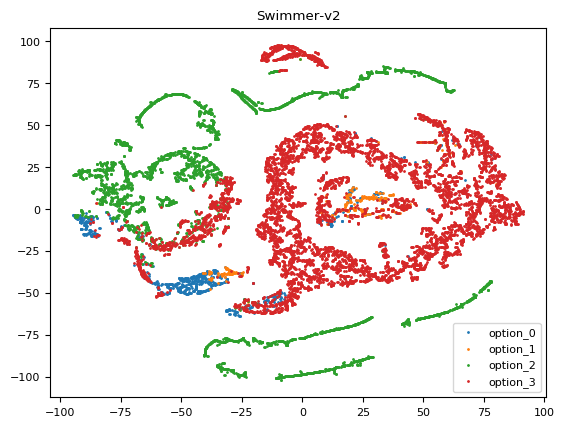}
        \label{fig:Swimmer_oc_tsne}
    }%
    \subfigure[\textit{Walker2D}]{
        \centering
        \includegraphics[width=0.33\linewidth]{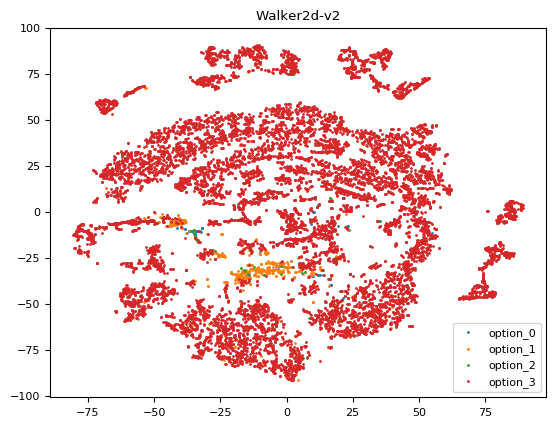}
        \label{fig:Walker2d_oc_tsne}
    }%

    \caption{t-SNE plot of the model trained with the option-critic architecture in MuJoCo Environment.}
    \label{fig:mujoco_oc_tsne}
\end{figure}
\phantom{blank}

\newpage
\begin{figure}[!ht]
    \centering
    \subfigure[\textit{HalfCheetah}]{
        \centering
        \includegraphics[width=0.33\linewidth]{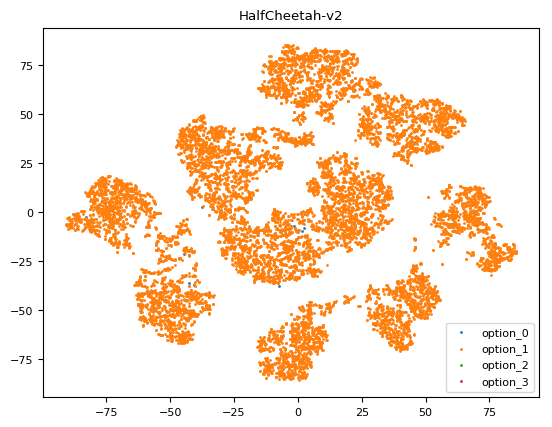}
        \label{fig:Halfcheetah_hd_tsne}
    }%
    \subfigure[\textit{Hopper}]{
        \centering
        \includegraphics[width=0.33\linewidth]{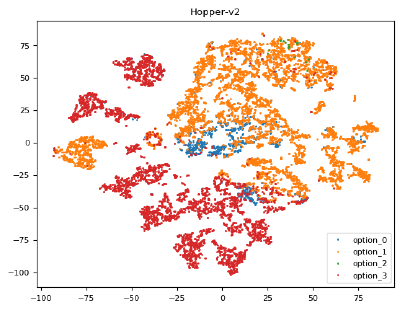}
        \label{fig:Hopper_hd_tsne}
    }%
    \vskip\baselineskip
    \subfigure[\textit{Swimmer}]{
        \centering
        \includegraphics[width=0.33\linewidth]{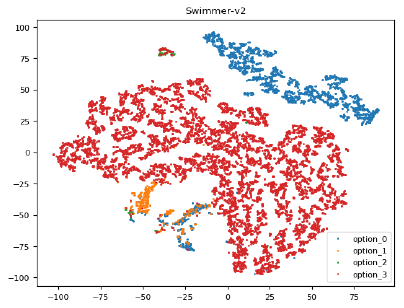}
        \label{fig:Swimmer_hd_tsne}
    }%
    \subfigure[\textit{Walker2D}]{
        \centering
        \includegraphics[width=0.33\linewidth]{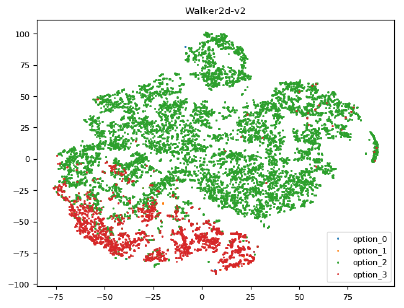}
        \label{fig:Walker2d_hd_tsne}
    }%
    \caption{t-SNE plot of the model trained with the hd-regularizer method in MuJoCo Environment.}
    \label{fig:mujoco_oc_tsne}
\end{figure}